%% file: aaai22.tex
\def\year{2022}\relax
\documentclass[letterpaper]{article} 
\usepackage{aaai22}  
\usepackage{times}  
\usepackage{helvet}  
\usepackage{courier}  
\usepackage[hyphens]{url}  
\usepackage{graphicx} 
\usepackage{natbib}  
\usepackage{caption} 
\DeclareCaptionStyle{ruled}{labelfont=normalfont,labelsep=colon,strut=off} 
\usepackage{algorithm}
\usepackage{algorithmic}
\usepackage{enumitem}

\usepackage{newfloat}
\usepackage{listings}
\floatstyle{ruled}
\newfloat{listing}{tb}{lst}{}
\floatname{listing}{Listing}
\title{Towards Fine-Grained Reasoning for Fake News Detection}

\author{
    Yiqiao Jin\textsuperscript{\rm 1}\thanks{Work done during an internship at Microsoft Research Asia.},
    Xiting Wang\textsuperscript{\rm 2}\footnote{Xiting Wang is the corresponding author.},
    Ruichao Yang\textsuperscript{\rm 3},
    Yizhou Sun\textsuperscript{\rm 1},
    Wei Wang\textsuperscript{\rm 1},
    Hao Liao\textsuperscript{\rm 4},
    Xing Xie\textsuperscript{\rm 2}
}
\affiliations{
    \textsuperscript{\rm 1} University of California, Los Angeles,
    \textsuperscript{\rm 2} Microsoft Research Asia,\\
    \textsuperscript{\rm 3} Hong Kong Baptist University,
    \textsuperscript{\rm 4} Shenzhen University \\
    ahren2040@g.ucla.edu,
    \{xitwan, xing.xie\}@microsoft.com, \\
	\{yzsun, weiwang\}@cs.ucla.edu, 
    csrcyang@comp.hkbu.edu.hk,
    haoliao@szu.edu.cn

}

\usepackage{bibentry}

\usepackage{xcolor}
\usepackage{amsmath}
\usepackage{amsfonts}
\usepackage{multirow}

\newcommand{\xiting}[1]{\textcolor{black}{#1}}
\newcommand{\yiq}[1]{\textcolor{black}{#1}}

\begin{document}

\maketitle

\begin{abstract}


The detection of fake news often requires sophisticated reasoning skills, such as logically combining information by considering word-level subtle clues. In this paper, we move towards fine-grained reasoning for fake news detection by better reflecting the logical processes of human thinking and enabling the modeling of subtle clues. In particular, we propose a fine-grained reasoning framework by following the human’s information-processing model, introduce a mutual-reinforcement-based method for incorporating human knowledge about which evidence is more important, and design a prior-aware bi-channel kernel graph network to model subtle differences between pieces of evidence. Extensive experiments show that our model outperforms the state-of-the-art methods and demonstrate the explainability of our approach.


\end{abstract}

\input{src/introduction}
\input{src/problem_statement}

\input{src/method}
\input{src/evaluation}
\input{src/related}
\input{src/conclusion}

\bibliography{aaai22}

\input{src/supplementary}


\end{document}

%% file: src/introduction.tex
\section{Introduction}


The emergence of social media has transformed the way users exchange information online.
People are no longer mere reviewers of information, but content creators and message spreaders.
Consequently, it has become much easier for fake news to spread on the Internet. 
Since fake news can obscure the truth, undermine people's belief, and cause serious social impact~\cite{brewer2013impact}, detecting fake news has become increasingly important for a healthy and clean network environment~\cite{shu2017fake}.
 
Recently, neural models have been proposed to detect fake news in a data-driven manner~\cite{pan2018content,dun2021kan}.
These works have shown the promise in
leveraging big data for fake news detection.
However, works that approach the task from the perspective of \emph{reasoning} are still lacking.
According to the literature on psychology, reasoning is the capability of \emph{consciously applying logic} for truth seeking~\cite{honderich2005oxford}, and is typically considered as a distinguishing capacity of \emph{humans}~\cite{mercier2017enigma}.
We observe that such ability is essential to improve the explainability and accuracy for fake news detection: 

\textbf{Explainability}.
Most existing works on fake news detection either do not provide explanations or enable explainability for a small part of the model (e.g., the attention layer). 
The major part of the model (e.g. the overall workflow) remains obscure to humans. 
This prevents humans to better understand and trust the model, or steer the model for performance refinement.

\begin{figure}[t]
    \centering
    \includegraphics[width=0.5\textwidth]{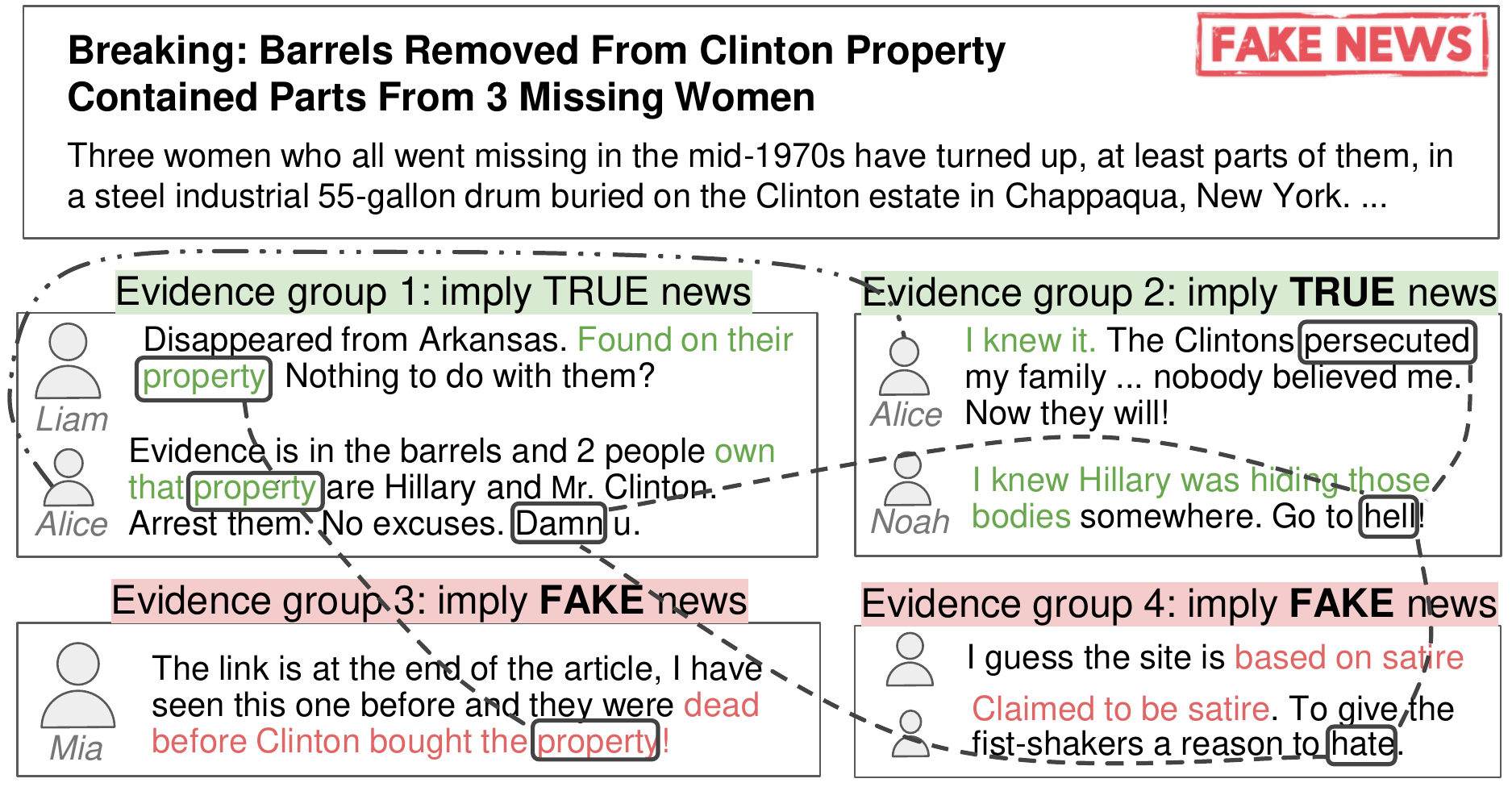}
    \vspace{-7mm}
    \caption{
        A motivating example of fine-grained reasoning for fake news detection.
    }
    \vspace{-7mm}
    \label{fig:intro-example}
\end{figure}

\textbf{Accuracy}. When humans reason about the authenticity of a news article, they have the ability to perform \emph{fine-grained} analysis for identifying subtle (e.g., word-level) clues, and can connect different types of clues (e.g., textual and social) to draw conclusions. 
An example is shown in Fig.~\ref{fig:intro-example}. 
Although the four groups of evidence are semantically dissimilar, humans can logically connect them in terms of subtle clues such as the word ``property", 
which leads to a much more confident conclusion about the article.
For example, reasoning in terms of ``property" reveals that the accusation of finding bodies in Clintons' property (evidence group 1) might be false,
since the women were dead before the Clintons bought the property (evidence group 3).
Reasoning with respect to ``hate" suggests that users in groups 1 and 2 might post false messages because they hate the Clintons.
The \xiting{overlap between users} in groups 1 and 2 further strengthens this suggestion.
Existing methods lack such capability of fine-grained reasoning:
they either do not model the interactions between different types of evidence or model them at a coarse-grained (e.g., sentence or post) level. 

\begin{figure*}[t]
    \centering
    \includegraphics[width=0.95\textwidth]{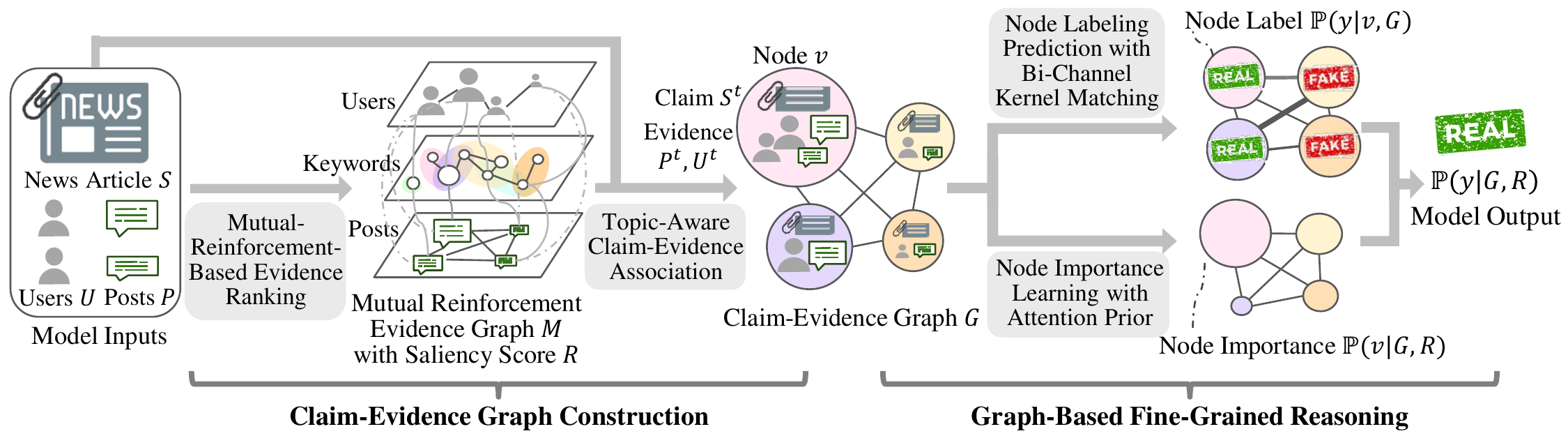}
     \vspace{-2mm}
    \caption{
Our proposed \emph{FinerFact} framework for fake news detection.
    } 
    \label{fig:model}
    \vspace{-4mm}
\end{figure*}

We aim to move towards using
deep reasoning for fake news detection.
The goal is to improve accuracy and explainability by 1) better reflecting the logical processes of human thinking and 2) enabling fine-grained modeling of subtle clues.
In particular, we study three research questions:
\begin{itemize}[nosep,leftmargin=1em,labelwidth=*,align=left]
    \item RQ1. Can the model be designed by following
    the human's information-processing model~\cite{lang2000limited}? 
    \item RQ2. Can human knowledge about which evidence (e.g., posts and users) is important be better incorporated? 
    \item RQ3. How does one achieve fine-grained modeling of different types of subtle clues?
\end{itemize}

Specifically, we make the following contributions.

    First, we design a \textbf{\underline{Fine}-grained \underline{r}easoning framework for \underline{Fa}ke news dete\underline{ct}ion} (\textbf{FinerFact}) by following the human's information-processing model (RQ1).
This enables us to detect fake news by better reflecting the logical processes of human thinking, which enhances interpretability and provides the basis for incorporating human knowledge.

Second, we propose a \textbf{mutual-reinforcement-based method} for evidence ranking, which enables us to better incorporate prior human knowledge about which types of evidence are the most important (RQ2). 

Finally, we design a \textbf{prior-aware bi-channel kernel graph network} to achieve fine-grained reasoning by modeling different types of subtle clues (RQ3). 
Our method improves accuracy, and 
provides explanations about the subtle clues identified, the most important claim-evidence groups, and the individual prediction scores given by each group.\looseness=-1

%% file: src/problem_statement.tex
\section{Methodology}

\subsection{Problem Definition}

\textbf{Input.} The inputs of our model are threefold. Each training sample consists of
1) a news article $S$ to be verified;
2) a collection of online posts $P$ for the news and the commenting/retweeting relationships between the posts; and 
3) the online users $U$ 
that publish the posts $P$. 

\noindent \textbf{Output.} 
The output of our model is the predicted label of the news, which can be fake $(y = 1)$ or real $(y = 0)$.

%% file: src/method.tex
\subsection{Fine-Grained Reasoning Framework}

We propose a \emph{\underline{Fine}-grained \underline{r}easoning framework for \underline{Fa}ke news dete\underline{ct}ion} 
by following the human's information-processing model~\cite{lang2000limited}.
Our framework consists of two major modules as shown in Fig.~\ref{fig:model}.

The first module, \textbf{claim-evidence graph construction}, corresponds to the \emph{storage} sub-process of the human's information-processing model, in which people select the most important pieces of information and build their in-between associations to store them in the memory.
As of fake news detection, 
it corresponds to the process in which people search for key information such as the major viewpoints, opinion leaders, and the most important posts,
which enables them to get an idea about the key claims and their associated evidence (e.g., supported posts and users).
This step is essential for filtering noise, organizing facts, and speeding up the fine-grained reasoning process at the later stage. 
It also enables us to incorporate human knowledge about which information is important.


The second module, \textbf{graph-based fine-grained reasoning}, corresponds to the \emph{retrieval} sub-process of the human's information-processing model, in which people reactivate specific pieces of information based on their associations for decision making.
In fake news detection, this module enables fine-grained modeling of evidential relations by considering subtle clues, such as the word ``property", ``hate", and the overlapping user in Fig.~\ref{fig:intro-example}.

By designing the framework based on the human's information-processing model, the overall workflow of our method resembles the logical processes used by humans, and most of the intermediate results are understandable by ordinary users. This provides a good basis for users to trust the model and steer it by integrating human knowledge.





\subsection{Claim-Evidence Graph Construction}

Our graph construction method contains two parts: 
1) \textbf{mutual-reinforcement-based evidence ranking}, 
which distinguishes important evidence from noise
by incorporating human knowledge
(filtering noise); 
and 2) \textbf{topic-aware claim-evidence association}, which 
follows journalists' knowledge about quality journalism~\cite{journalismbook} to extract
the key claims and associate them with the corresponding evidence (organizing facts).


\subsubsection{Mutual-Reinforcement-Based Evidence Ranking}

As a news article propagates on the Internet, it leaves many traces, e.g., posts about the news and users who support the news, which can all be considered as evidence for verifying the news.
Such evidence typically has a large scale, 
and performing fine-grained reasoning by considering all evidence is quite difficult, if not impossible, due to limited GPU memory.
To efficiently and accurately identify the most valuable evidence in an interpretable way, we propose ranking 
the evidence by integrating human knowledge.
The human knowledge can be divided into two types based on whether it considers the inherent \textbf{attributes}~\cite{lampos2014predicting} of the evidence or its \textbf{topological}~\cite{pasquinelli2009google} information.
We observe that these two types of knowledge can be integrated in one framework and computed efficiently by using the mutual reinforcement mechanism~\cite{duan2012twitter}.
Specifically, our mutual-reinforcement-based evidence ranking consists of the following three steps.

\emph{\underline{Step 1: Attribute saliency computation.}}
We compute the attribute saliency $E$ based on the human knowledge summarized from the current literature.
In particular, the attribute saliency $e_{u_i}$ for a \textbf{user} $u_i$ is computed by using the user impact index~\cite{lampos2014predicting}:
\begin{eqnarray}
    & e_{u_i}=\textrm{ln}\left(\frac{(\phi_i^{I}+\theta_U)^{2}(\phi_i^{L}+\theta_U)}{\phi_i^{O}+\theta_U}\right)
\label{eqn:user_saliency}
\end{eqnarray}
where $\phi_i^{ I }$, $\phi_i^{ O }$, $\phi_i^{ L }$ are $u_i$'s follower count, friend count, and listed count, and $\theta_U$ is a smoothing constant added to ensure that the minimum saliency score is always positive.
The attribute saliency $e_{p_i}$ of each \textbf{post} $p_i$ is computed based on its number of retweets $C_i$, considering that retweets rarely provide extra textual information and are usually regarded as votes to the original post \cite{chang2013towards}.
Specifically, $e_{p_i} = \textrm{ln}(C_{i}+1) + \theta_P$, where $\theta_P$ is a smoothing constant. 
The attribute saliency $e_{k_i}$ of each \textbf{keyword} $k_i$ is set to
$\textrm{ln}(\mathrm{freq}(k_i)+1) + \theta_K$, where $\mathrm{freq}(k_i)$ is the term frequency of $k_i$ with respect to the news article and the posts, and $\theta_K$ is a smoothing constant \cite{rousseau2013graph}.

\emph{\underline{Step 2: Mutual reinforcement evidence graph building.}}
As shown in Fig.~\ref{fig:model}, we build the mutual reinforcement evidence graph $M$ so that it encodes both the relations within posts, users, or keywords, and the relations between them to enable the effective modeling of topological saliency.
Mathematically, $M=\{A_{xy}|x,y\in\{P,U,K\}\}$ is a three-layer graph, where subscript indices $P$, $U$, $K$ denote posts, users, and keywords, respectively, and $A_{xy}$ is an affinity matrix that represents relations between items.
We design the graph based on two considerations:
1) $M$ should effectively encode diverse types of social interactions 
and 2) edges in $M$ should reveal the mutual influence between items in terms of saliency.
For example, constructing edges between users and their posts means that if a user is important, then his/her posts are important, and vice versa. 
Such human knowledge about which items should influence each other in terms of saliency can be effectively incorporated into the edge design.
Based on these considerations, we construct $M$ by using the cosine similarity between the term frequency vectors of the posts ($A_{PP}$), the commenting relationships between users ($A_{UU}$), the co-occurrence relationships between keywords ($A_{KK}$), the mentioning relationships that link a keyword to all the posts and users that mention it ($A_{KP}$, $A_{KU}$), and the authoring relationships that link a user to all the posts that s/he has published ($A_{UP}$).  
More details about the construction of $M$ and how human knowledge is used are given in the supplement.\looseness=-1



\emph{\underline{Step 3: Iterative saliency propagation.}}
We then compute the saliency $R$ based on the mutual reinforcement mechanism~\cite{duan2012twitter}.
In particular, we treat the attribute saliency $E$ as a prior and integrate it with the saliency propagation process on $M$:

\begin{equation}
R^{(i+1)} = d\tilde{A}R^{(i)}+ (1-d)E
\label{eq:mrg}
\end{equation}
\begin{equation*}
\resizebox{0.48\textwidth}{!}{
$\hat{A} = \begin{bmatrix}\beta_{P P} A_{P P} & \beta_{K P} A_{K P} & \beta_{U P} A_{U P} \\ \beta_{P K} A_{P K} & \beta_{K K} A_{K K} & \beta_{U K} A_{U K} \\ \beta_{P U} A_{P U} & \beta_{K U} A_{K U} & \beta_{U U} A_{U U}\end{bmatrix}, 
R^{(i)} = \begin{bmatrix} R_P^{(i)} \\ R_K^{(i)} \\ R_U^{(i)}\end{bmatrix}, 
E = \begin{bmatrix} E_P \\ E_K \\ E_U \end{bmatrix}$
}
\end{equation*}
where $R^{(i)}$ is the joint ranking score vector in the $i$-th iteration, $\tilde{A}$ is the normalized affinity matrix derived from $\hat{A}$, and $\beta_{xy}$ is a balancing weight to adjust the interaction strength among posts, keywords, and users.

\emph{\underline{Interpretability and efficiency.}}
The designed method is highly interpretable and steerable, since each type of saliency can be easily explained to and controlled by human users.
According to Eq.~(\ref{eq:mrg}), a piece of evidence (e.g., a supported user) is important if it has a large attribute saliency (e.g., has many followers) and is connected with other salient evidence (e.g., writes a salient post).
Other types of human knowledge can also be easily integrated by slightly modifying the equation. 
Our method is also efficient.
In practice, ranking 240,000 posts, users, and keywords takes 620 seconds.
Without evidence ranking, performing fine-grained reasoning on the same data causes the out of memory issue on NVIDIA Tesla V100.

\subsubsection{Topic-Aware Claim-Evidence Association}
Given the saliency scores $R$, a straightforward way for constructing the claim-evidence graph is to select the pieces of evidence with the largest saliency scores.
However, this method may easily obscure the  truth by focusing only on one aspect of the story.
For example, the news article in Fig.~\ref{fig:intro-example} may be dominated by posts related to evidence group 1, which are posted deliberately by users who hate the Clintons.
To disclose the truth, we need to observe all four evidence groups closely.
This echoes the journalists' knowledge about quality journalism~\cite{journalismbook}, which states that a high-quality news article should cover multiple viewpoints to reveal two or more sides of the story.
Motivated by this insight, we propose a topic-aware method, which consists of the following steps. 

\underline{\emph{Step 1: Topic modeling.}}
A typical solution to mine major viewpoints in a text corpus is topic modeling.
In the scenario of fake news detection, the text corpus consists of each post and news sentence.
We then utilize LDA~\cite{blei2003latent} to mine the topics,
which summarizes the main viewpoints and serve as a bridge between the claims in the news sentences and the keywords in the evidence graph $M$:
\begin{itemize}[nosep,leftmargin=1em,labelwidth=*,align=left]
    \item Each topic $t$ is represented by a distribution of keywords $\mathbb{P}(k_i|t)$ in $M$. For each topic $t$, 
    its top $N_K$ keywords $K^t$ are the ones that have the largest $\mathbb{P}(k_i|t)$.
    \item Each news sentence $s_i$ is represented by a distribution of topics $\mathbb{P}(t|s_i)$. Given a topic $t$, 
    we can extract its top $N_S$ sentences $S^t$ that have the largest $\mathbb{P}(t|s_j^t)$.
\end{itemize}



\underline{\emph{Step 2: Key claim and evidence extraction.}}
To extract the key claims and their associated evidence based on the topics, we first select the top $N_T$ topics with the maximum 
aggregate saliency score 
$r_t=\sum_{k_i\in K^t} \mathbb{P}(k_i|t) r_{k_i}$, 
where $K^t$ is the set of top keywords extracted in step 1, and  $r_{k_i}$ is the saliency score computed by using mutual-reinforcement-based evidence ranking. 
The topics that are selected in this way not only cover major viewpoints of the article, but are also related with the most salient evidence in $M$.

For each selected topic $t$, its corresponding claims are the top news sentences $S^t$ extracted in step 1.
The key evidence consists of two parts.
The first part is the key posts in $t$.
Given a set of posts that are connected to the keyword set $K^t$,
we normalize the saliency score of each post in the set to obtain a probability distribution, and then sample a set $P^t$ with $N_P$ posts according to the distribution.
Similarly, we sample a set $U^t$ with $N_U$ users that are relevant with $t$, and treat $U^t$ as the second part of the evidence.
Keywords are not considered as evidence because they are less useful for fine-grained reasoning when considered without the context.
To model keywords more effectively, we treat the posts as ordered sequences of keywords in fine-grained reasoning.

\underline{\emph{Step 3: Graph construction.}}
Finally, we build a claim-evidence graph $G$ as shown in Fig.~\ref{fig:model}.
In $G$, each node $v$ is a tuple $(t, S^t, P^t, U^t)$ that corresponds to a selected topic $t$, where $S^t$ refers to the key claims and $P^t$, $U^t$ form an evidence group shown in Fig.~\ref{fig:intro-example}.
Given all nodes $V$, a straightforward approach to construct $G$ is to build edges between two nodes if \xiting{the percentage of overlapping words} is larger than a threshold~\cite{zhong2020reasoning}.
However, this method may easily overlook important subtle clues, because 1) it is difficult to find an appropriate global threshold and 2) topics (or evidence groups) may be connected logically with different words (e.g., `damn", ``hate", and ``hell" in Fig.~\ref{fig:intro-example}).
Based on this observation, we choose to build a fully connected graph and let the fine-grained reasoning module to decide whether subtle clues exist for a set of topics.
\subsection{Fine-Grained Graph-based Reasoning}

After constructing the claim-evidence graph $G$, we model subtle clues and effectively leverage them for prediction through fine-grained graph-based reasoning. 
Our method is based on the Kernel Graph Attention Network (KGAT)~\cite{liu2020fine}.
We choose this method because it can effectively model subtle differences between statements and propagate the learned information on the graph.
However, KGAT cannot be directly applied on our claim-evidence graph, because it handles only textual inputs and cannot integrate the learned saliency $R$, which incorporates human knowledge about which evidence is important.  
To solve these issues, we propose a Prior-Aware Bi-Channel KGAT that extends KGAT to 1) simultaneously model subtle clues from both textual (posts) and social (users) inputs with two \xiting{connected} channels; and 2) integrating existing knowledge about important evidence with attention priors.
Mathematically, the final prediction $\mathbb{P}(y |G, R)$ is obtained by combining individual node-level predictions:
\begin{eqnarray}
\label{eqn:prediction}
\centering
\mathbb{P} (y \mid G, R)= \sum_{v\in G} \underbrace{\mathbb{P}(y \mid v, G)}_{\substack{\text{Node label}\\ \text{prediction}}} \underbrace{\mathbb{P}(v \mid G, R)}_{\substack{\text{Node importance}\\ \text{learning}}}
\end{eqnarray}

This formulation provides explainability for
individual prediction scores that each node (or claim-evidence group) gives and importance of the nodes for the final prediction.
Since we do not have node-level annotations for label prediction or importance,
we learn them automatically with:
\begin{itemize}[nosep,leftmargin=1em,labelwidth=*,align=left]
    \item \textbf{Node label prediction with bi-channel kernel matching}, which accurately computes $\mathbb{P}(y | v, G)$ by integrating different types of subtle clues from the whole graph;
    \item \textbf{Node importance learning with attention priors}, which effectively models $\mathbb{P}(v | G, R)$ by integrating the evidence saliency $R$ as attention priors.  
\end{itemize}

\subsubsection{Node Label Prediction with Bi-Channel Kernel Matching}
Given a node $v$, we predict the individual label it gives by aggregating subtle clues from the whole graph.
To reason about subtle clues that are provided by both the textual and social (user) inputs, 
we design two interconnected channels.
We will first introduce how each channel is designed when they are modeled \emph{independently}, and then illustrate how the channels can be \emph{fused} for prediction. 

\emph{\underline{Text-based reasoning channel.}}
We first derive an initial textual representation for node $v=(t, S^t, P^t, U^t)$ by 
concatenating the claims in $S^t$ and evidential posts in $P^t$ with token ``[SEP]''. The concatenated string is then encoded by using BERT~\cite{devlin2019bert}:
\begin{eqnarray}
[\mathbf{h}_v^0, \mathbf{h}_v^1, ... \mathbf{h}_v^{N_v}] = \textrm{BERT}(S^t \oplus P^t), \ \ \ \ \mathbf{z}_v = \mathbf{h}_v^0
\end{eqnarray}
where $\mathbf{h}_v^i$ denotes the BERT embedding for the $i$-th token.
$\mathbf{z}_v$, which corresponds to the embedding of the ``[CLS]'' token, is considered as the initial textual representation for $v$.

The fine-grained match features between nodes $v$ and $q$ can then be extracted by constructing a token-level translation matrix $L_{q, v}$.
In $L_{q, v}$, each entry $l_{q, v}^{i,j}=\textrm{cos}(\mathbf{h}_q^i, \mathbf{h}_v^j)$ is the cosine similarity between their token representations. 
For each token $i$ in node $q$, we use $\Upsilon$ kernels to extract the kernel match features between the token and its neighbor $v$:
\begin{eqnarray}
\label{eqn:kernel_match}
\indent & \Psi_{\tau}(L_{q, v}^{i})=\log \sum_{j} \exp (-\frac{(l_{q, v}^{i, j}-\mu_{\tau})^{2}}{2 \sigma_{\tau}^{2}})  \\ 
\indent & \vec{\Psi}(L_{q, v}^i) = \{\Psi_1(L_{q, v}^{i}), \ldots , \Psi_{\Upsilon}(L_{q, v}^{i})\} 
\end{eqnarray}
\noindent Each $\Psi_{\tau}$ is a Gaussian kernel that concentrates on the region defined by the mean similarity $\mu_{\tau}$ and the standard deviation $\sigma_\tau$.
The kernel match feature $\vec{\Psi}$, which consists of $\Upsilon$ kernels, summarizes how similar the $i$-th token in $q$ is to all tokens in $v$ at different levels. 
Such soft-TF match features have been shown effective for fact verification~\cite{liu2020fine}.
In our framework, they help identify subtle clues by comparing different claim-evidence groups (nodes).
For example, in Fig.~\ref{fig:intro-example}, the match features can help identify ``hate" in group 4 (node $q$)  by comparing it with all words in group 2 (node $v$).
This neighbor-aware token selection is achieved by computing an attention score based on $\vec{\Psi}$:
\begin{eqnarray}
 &\alpha_{q, v}^i = \textrm{softmax}_i(W_1 \vec{\Psi}(L_{q, v}^i) + b_1)
 \label{eqn:token_select}
\end{eqnarray}
We can then compute the content to propagate from $q$ to $v$:
\begin{eqnarray}
 &\hat{\mathbf{z}}_{q, v} = \sum_{i} \alpha_{q, v}^i \mathbf{h}_q^i
 \label{eqn:token_select_2}
\end{eqnarray}
Note that by setting $\sigma_\tau$ to $\infty$, Eq.~(\ref{eqn:token_select_2}) degenerates to mean pooling, which assigns an equal weight to all tokens.

Given $\hat{\mathbf{z}}_{q, v}$ that contains the information to be propagated to $v$, we derive a final textual representation $\kappa_v$ for $v$ by attentively aggregating $\hat{\mathbf{z}}_{q, v}$ from all $q\in G$: 
\begin{eqnarray}
\label{eqn:kgat_textual_1}
\indent & \kappa_v = (\sum_{q\in G} \gamma_{q, v} \cdot \mathbf{\hat{z}}_{q, v}) \oplus \mathbf{z}_v  \\ 
\label{eqn:kgat_textual_2}
\indent & \gamma_{q, v} =  \mathrm{softmax}_q(\textrm{MLP}(\mathbf{\hat{z}}_{q, v} \oplus \mathbf{z}_{v})) 
\raisetag{30pt}
\end{eqnarray}

The kernel-based textual representation $\kappa_v$ aggregates fine-grained, token-level subtle clues from the whole graph, and can be used to reason about the authenticity of the news from the perspective of node $v$.
Next, we introduce how to derive kernel-based user representations with a user channel, and how these two channels can be fused for final prediction.

\emph{\underline{User-based reasoning channel.}}
The initial user representation $\mathbf{x}_v$ for node $v=(t, S^t, P^t, U^t)$ is derived by applying a graph neural network, APPNP~\cite{klicpera2018predict}, on the mutual reinforcement evidence graph $M$.
We choose APPNP because its efficient message-passing scheme enables it to scale to large graphs with hundreds of thousands of nodes.
Specifically, for each user in $U^t$, we first obtain its feature embedding by using a look up layer, which encodes the main user attributes including the user's follower count, friend count, listed count, favourite count, status count, the number of words in the self description, as well as the account status about whether the user is verified or geo-enabled.
We then use the message passing scheme of APPNP to aggregate the feature embeddings from the neighbor users in $M$.
This results in an initial user representation $\mathbf{u}_v^i$ for each user, and max pooling is used to derive the initial user presentation $\mathbf{x}_v$ for node $v$:
\begin{eqnarray}
[\mathbf{u}_v^0, ... \mathbf{u}_v^{\tilde{N}_v}] = \textrm{APPNP}(M), \ \  
\mathbf{x}_v = \text{maxpool}(\mathbf{u}_v^0, ... \mathbf{u}_v^{\tilde{N}_v})
\end{eqnarray}
We can then derive the kernel-based user representation $\tilde{\kappa}_v$ by using a  kernel attention mechanism similar with that in the text-based reasoning channel:
\begin{eqnarray}
\label{eqn:kgat_user_1}
\indent & \rho_{q, v}^i = \mathrm{softmax}_i(W_2 \vec{\Psi}(\tilde{L}_{q, v}^i) + b_2)  \\ 
\indent & \mathbf{\hat{x}}_{q, v} = \sum_{i} \rho_{q, v}^i \mathbf{u}_{q}^i  \\
\label{eqn:kgat_user_3}
& \lambda_{q, v} =  \mathrm{softmax}_q(\textrm{MLP}(\mathbf{\hat{x}}_{q, v} \oplus \mathbf{x}_{v}))  \\
& \tilde{\kappa}_v = (\sum_{q} \lambda_{q, v} \cdot \mathbf{\hat{x}}_{q, v}) \oplus \mathbf{x}_v \label{eqn:kgat_user_interact}
\end{eqnarray}
\noindent where $\tilde{L}_{q, v}$ is a user-level translation matrix 
in which each entry is a cosine similarity score between two initial user representations. 
This formulation of $\tilde{\kappa}_v$ allows us to reason about the final prediction by considering user-based subtle clues, e.g., the overlap between users in Fig.~\ref{fig:intro-example}.


\underline{\emph{Channel fusion.}}
We fuse the channels to better integrate information from the textual and social inputs. 
To this end, we first refine the node-level attention scores by aggregating the textual and user representations.
Specifically, we replace  $\gamma_{q, v}$ and $\lambda_{q, v}$ in Eqs.~(\ref{eqn:kgat_textual_2}) and (\ref{eqn:kgat_user_3}) with $\mathrm{softmax}_q(\textrm{MLP}(\mathbf{\hat{z}}_{q, v} \oplus \mathbf{z}_{v} \oplus \mathbf{\hat{x}}_{q, v} \oplus \mathbf{x}_{v}))$.
This allows us to combine both textual and social clues when reasoning about one node based on another node.
For example, in Fig.~\ref{fig:intro-example}, we may consider both words related to ``hate" and the overlapping user when reasoning about evidence group 1 from the perspective of evidence group 2.




We then fuse the kernel-based textual and user representations to predict the label that node $v$ gives:  
\begin{eqnarray}
\label{eqn:channel_fuse}
\begin{split}
& \mathbb{P}(y \mid v, G) = \textrm{sigmoid}_{v}(W_5(\kappa_{v}) + W_6(\mathbf{\tilde{\kappa}}_v) + b_5)
\end{split}
\end{eqnarray}

\begin{table*}[]
\centering
\resizebox{0.88\textwidth}{!}{
\begin{tabular}{cc|ccccc|ccccc} 
\hline
                                           &           &                 &                 & \multicolumn{2}{l}{PolitiFact}    &                 &                 &                 & \multicolumn{2}{l}{GossipCop}     &                  \\ 
\cline{3-12}
\begin{tabular}[c]{@{}l@{}}\\\end{tabular} &           & Pre             & Rec             & F1              & Acc             & AUC             & Pre             & Rec             & F1              & Acc             & AUC              \\ 
\hline
                                           & SVM       & 0.7460          & 0.6826          & 0.6466          & 0.6694          & 0.6826          & 0.7493          & 0.6254          & 0.5955          & 0.6643          & 0.6253           \\
\multirow{2}{*}{\textbf{G1}}         & RFC       & 0.7470          & 0.7361          & 0.7362          & 0.7406          & 0.8074          & 0.7015          & 0.6707          & 0.6691          & 0.6918          & 0.7389           \\
                                           & DTC       & 0.7476          & 0.7454          & 0.7450          & 0.7486          & 0.7454          & 0.6921          & 0.6922          & 0.6919          & 0.6959          & 0.6929           \\ 
                                           & GRU-2     & 0.7083          & 0.7048          & 0.7041          & 0.7109          & 0.7896          & 0.7176          & 0.7079          & 0.7079          & 0.718           & 0.7516           \\
\hline
                                           & B-TransE  & 0.7739          & 0.7658          & 0.7641          & 0.7694          & 0.8340          & 0.7369          & 0.7330          & 0.7340          & 0.7394          & 0.7995           \\
\multirow{2}{*}{\textbf{G2}}  & KCNN      & 0.7852          & 0.7824          & 0.7804          & 0.7827          & 0.8488          & 0.7483          & 0.7422          & 0.7433          & 0.7491          & 0.8125           \\
                                           & GCAN      & 0.7945          & 0.8417          & 0.8345          & 0.8083          & 0.7992          & 0.7506          & 0.7574          & 0.7709          & 0.7439          & 0.8031           \\
                                           & KAN       & 0.8687          & 0.8499          & 0.8539          & 0.8586          & 0.9197          & 0.7764          & 0.7696          & 0.7713          & 0.7766          & 0.8435           \\ 
\hline
\multirow{2}{*}{\textbf{Ours}}             & FinerFact & \textbf{0.9196} & \textbf{0.9037} & \textbf{0.9172} & \textbf{0.9092} & \textbf{0.9384} & \textbf{0.8615} & \textbf{0.8779} & \textbf{0.8685} & \textbf{0.8320} & \textbf{0.8637}  \\
                                           & Impv.     & +5.1\%          & +5.4\%          & +6.3\%          & +5.1\%          & +1.9\%          & +8.5\%          & +10.8\%         & +9.7\%          & +5.5\%          & +2.0\%           \\
\hline
\end{tabular}
}
\caption{Performance comparison of \emph{FinerFact} w.r.t. baselines. The best results are highlighted in \textbf{bold}.
}
\vspace{-1mm}
\label{table:perf}
\end{table*}

\subsubsection{Node Importance Learning with Attention Priors}
Node importance decides which topic (node) should be considered more in detecting fake news.
To better characterize the relative importance of each node with regard to the predicted label, we learn the probability $\mathbb{P}(v \mid G, R)$ by jointly consider its claims, evidence, and the evidence saliency $R$:
\begin{eqnarray}
\label{eqn:node_importance}
\centering
& \mathbb{P}(v \mid G, R) = \textrm{softmax}_{v\in G}(\varphi(v) + \delta(v,R) + b_6)  \\ 
& \varphi(v) = W_7 \left[ \text{average}_i ( \vec{\Psi}(\hat{L}_{S^t, P^t}^{i}))\right]  \\
& \delta(v,R) = W_8 R_{P^t} + W_9 R_{U^t} + W_{10} R_{K^t} 
\end{eqnarray}
where $\varphi(v)$ is the node ranking feature learned by comparing the claims with the evidence, and $\delta(v,R)$ is the attention prior used to encode the previously learned saliency score $R$, which embeds human knowledge about evidence significance. 
More specifically, $\varphi(v)$ is derived by using the kernel match feature  $\vec{\Psi}(\hat{L}_{S^t, P^t}^{i})$, where $\hat{L}_{S^t, P^t}$ is a token-level translation matrix that measures the cosine similarities between tokens in the claims $S^t$ and tokens in the supported posts $P^t$.
The attention prior $\delta(v,R)$ is learned by combining the saliency scores $R_{P^t}$, $R_{U^t}$, and $R_{K^t}$, which correspond to the top posts, users, and keywords that are the most relevant with the topic $t$ of the node $v$.
$W_8, W_9, W_{10}$ are \xiting{non-negative weight vectors} that enable us to re-weight each piece of evidence during fine-grained reasoning.

\subsubsection{Joint Optimization}
Let $\|\Theta\|$ be the $L 2$ norm of all model parameters. For each news article $S$, we compute the loss
\begin{equation}
\mathcal{L}_S = - y^* \log \left(\hat{p} \right) + (1-y^*) \log \left(1- \hat{p}\right)  +\lambda_{reg}\| \Theta \|^{2}
\end{equation}
where $y^*$ is its the ground-truth label, 
$\hat{p}=\mathbb{P}(y|G,R)$ is the probability learned based on Eq.~(\ref{eqn:prediction}), 
and $\lambda_{reg}$ is the regularization coefficient. 
The parameters are then optimized jointly by minimizing $\sum_{S\in\mathcal{N}}\mathcal{L}_S$, where $\mathcal{N}$ consists of all the news articles in the training set.

  

%% file: src/evaluation.tex
\section{Experiment}

%

\subsection{Experimental Setup}

\subsubsection{Dataset}
To evaluate the performance of \emph{FinerFact}, we conduct experiments on two benchmark datasets, PolitiFact and GossipCop~\cite{shu2020fakenewsnet},
which contain 815 and 7,612 news articles, and the social context information about the news, their labels provided by journalists and domain experts.
\xiting{We follow~\cite{dun2021kan} to preprocess the data and conduct experiments}.
More details about dataset and the preprocessing steps are given in the supplement.
\subsubsection{Baselines}

We compare our \emph{FinerFact} method with eight baselines, which can be divided into two groups:

 The first group (\textbf{G1}) is content-based methods, which leverage the textual or visual content of the news for fake news detection.
    G1 contains four baselines:  
    \textbf{SVM} ~\cite{yang2012automatic},
    \textbf{GRU-2}~\cite{ma2016detecting},
    \textbf{RFC}~\cite{kwon2013prominent}, and \textbf{DTC}~\cite{castillo2011information}.\looseness=-2

The second group (\textbf{G2}) consists of knowledge-aware methods that detect fake news by leveraging auxiliary knowledge such as knowledge graphs and social knowledge about the online posts.
This group includes four methods:   
 \textbf{B-TransE}~\cite{pan2018content},
 \textbf{KCNN}~\cite{wang2018dkn},
 \textbf{GCAN}~\cite{lu2020gcan} and
 \textbf{KAN}~\cite{dun2021kan}.


\subsubsection{Evaluation Criteria}
Our evaluation criteria include
Precision (\textbf{Pre}), Recall (\textbf{Rec}), the \textbf{F1} score, Accuracy (\textbf{Acc}), and Area Under the ROC curve (\textbf{AUC}). We conduct 5-fold cross validation 
and the average performance is reported.

\begin{figure}
  \centering
  \includegraphics[scale=0.27]{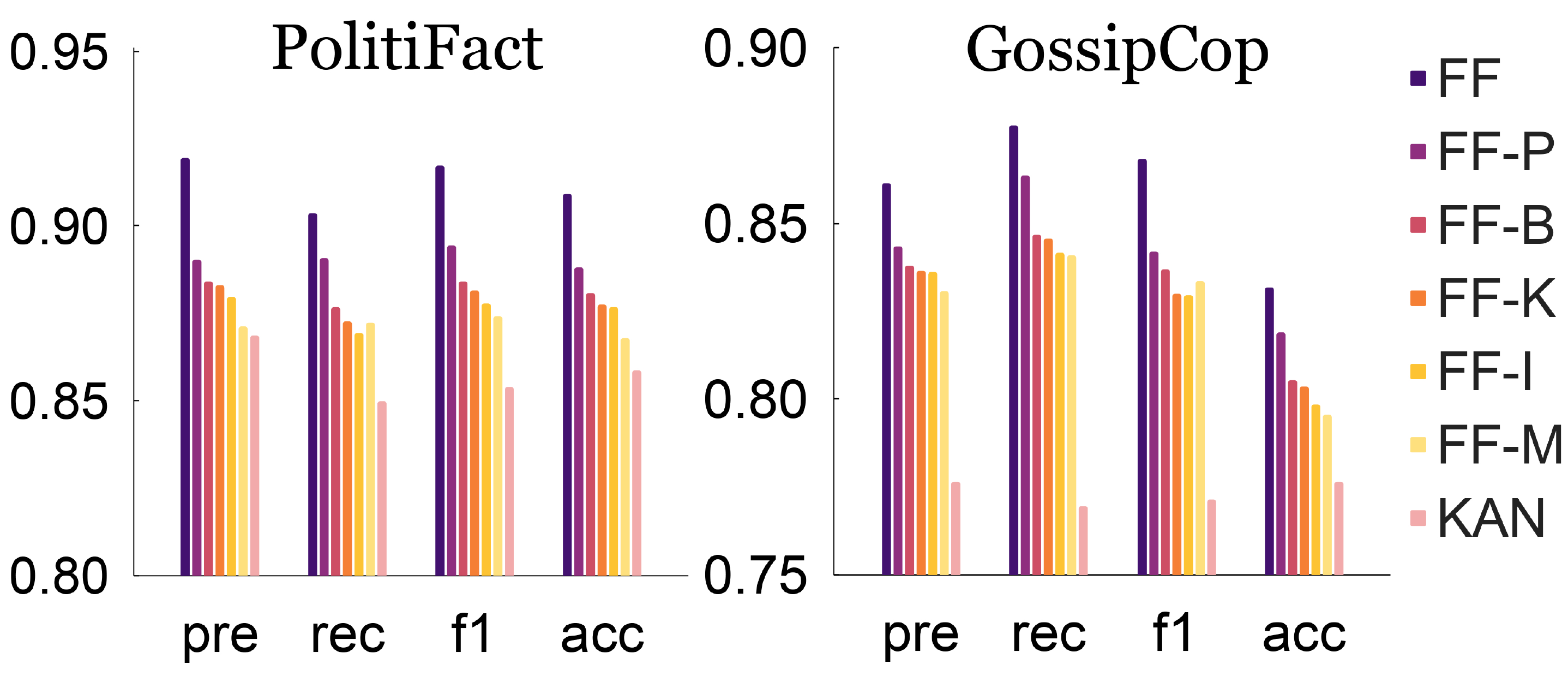}
  \vspace{-2mm}
  \caption{
  Results of the ablation study. 
  }
  \label{fig:ablation}
  \vspace{-2mm}
\end{figure}

\subsubsection{Implementation Details}
To choose the number of topics, we conducted a grid search within the range [2, 10] and picked the number that results in the smallest perplexity. BERT is fine-tuned during training.
A more comprehensive description about the implementation details, experimental setup, and evaluation for topic quality are in the supplement.


\subsection{Overall Performance}

Table \ref{table:perf} compares our method \emph{FinerFact} with the baselines.
As shown in the table, \emph{FinerFact} consistently outperforms the baseline in both datasets.
For example, \emph{FinerFact} performs better than the most competitive baseline \emph{KAN} by 6.3\%, 5.1\% on PolitiFact and 9.7\%, 5.5\% on GossipCop, respectively, in terms of the F1-score and  accuracy.
This demonstrates the effectiveness of our fine-grained reasoning framework, which enables the model to make predictions by identifying and connecting different types of subtle clues. 
Meanwhile, comparing with \emph{GCAN}, which models the interactions between the textual content and social information with co-attention, \emph{FinerFact} increases F1 by 8.3\%, 9.8\% on the two datasets. This implies that our kernel-attention-based approach can better model the interactions between news articles and evidence. 
We also observe that methods that incorporate external knowledge (\textbf{G2}) generally perform better than content-based methods (\textbf{G1}).
This illustrates the usefulness of external knowledge in fake news detection.

\subsection{Ablation Study and Sensitivity Analysis}

\begin{figure}[t]
  \centering
  \includegraphics[scale=0.27]{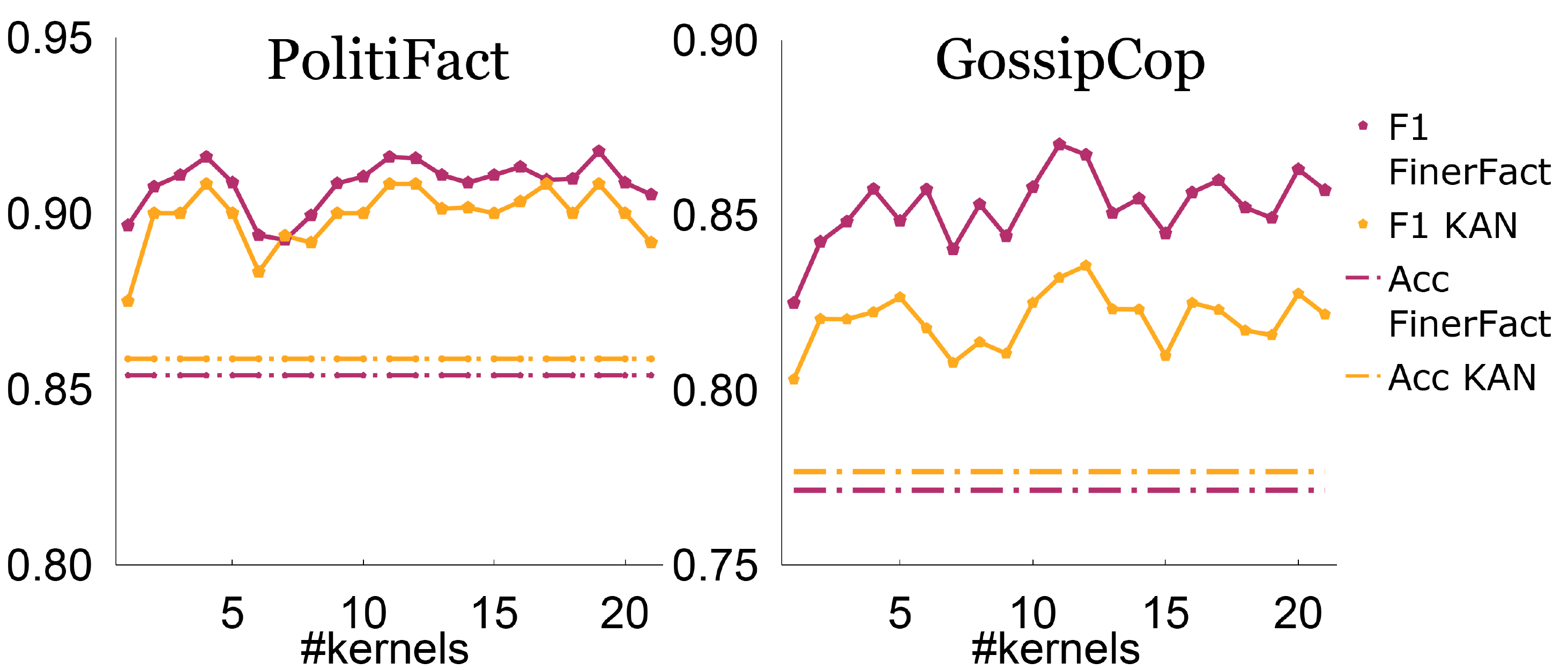}
  \caption{
  Sensitivity analysis w.r.t. the number of kernels.
  }
  \vspace{-2mm}
  \label{fig:sensitivity}
\end{figure}

\begin{figure*}
  \includegraphics[height=5cm]{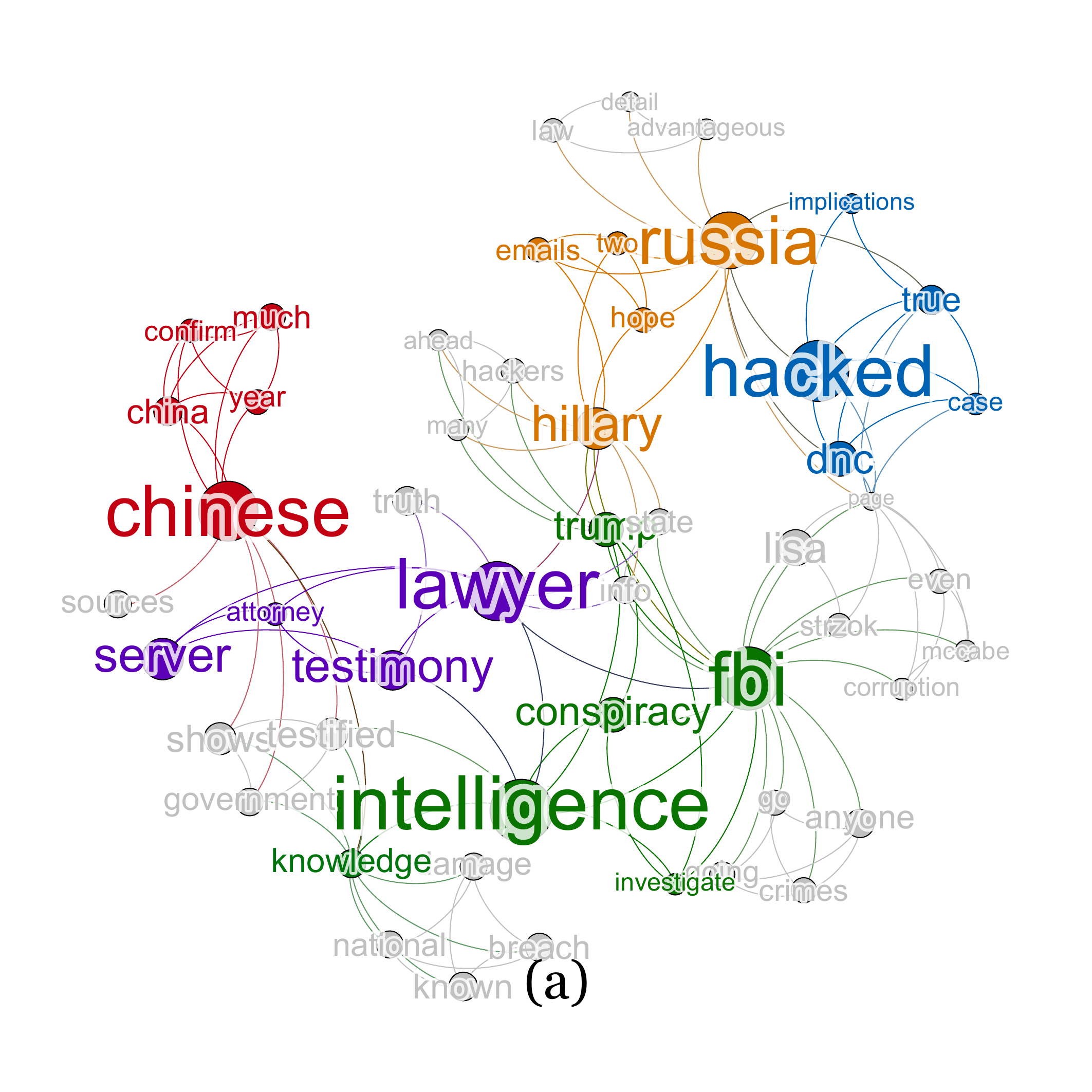}
  \includegraphics[height=5cm]{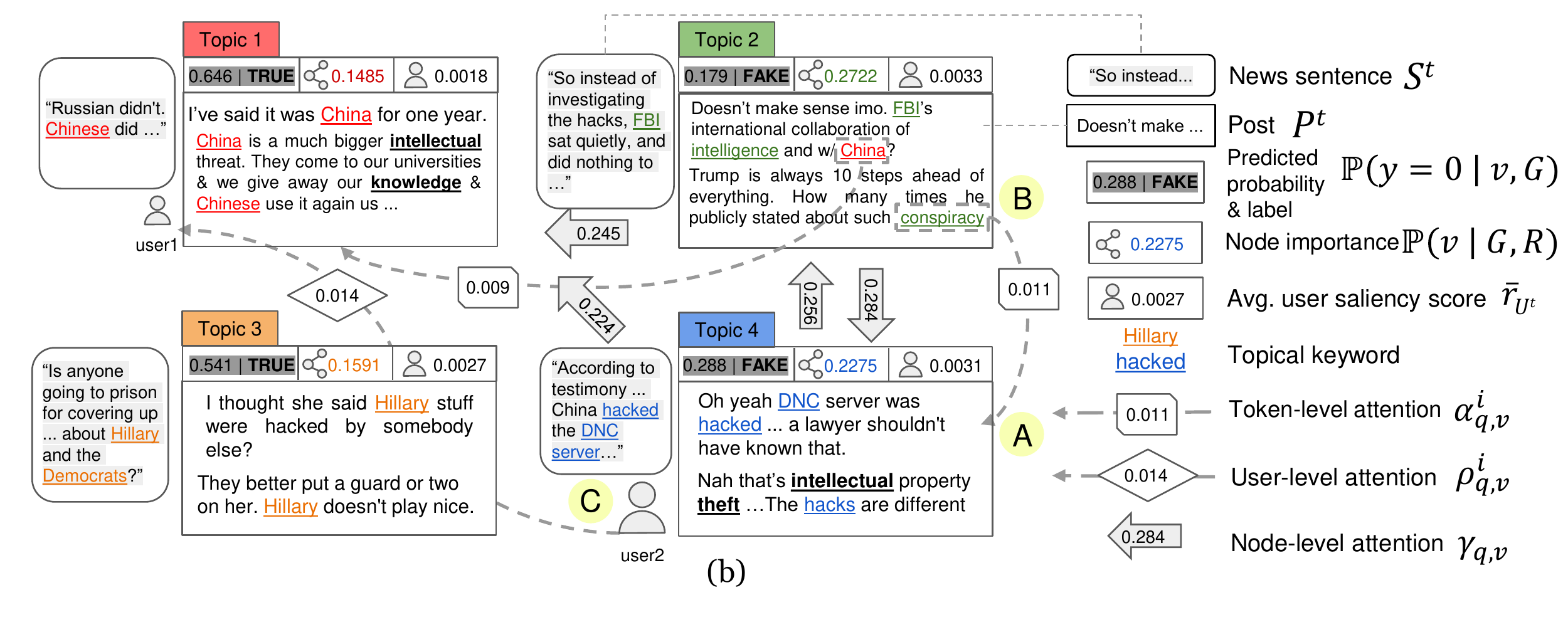}
  \vspace{-5mm}
  \caption{
  Reasoning with \emph{FinerFact}:
  (a) the keyword layer of the mutual reinforcement graph $M$, with saliency $R$ indicated by the font size; (b) fine-grained reasoning over the claim-evidence graph $G$.
  Each color encodes a topic.
  }
  \vspace{-1mm}
  \label{fig:case_study}
\end{figure*}

We conduct the ablation study by implementing five variants of our method:
1) \textbf{FF-P} removes the attention prior $\delta(v,R)$ when learning node importance;
2) \textbf{FF-B} eliminates bi-channel reasoning by removing the user-based reasoning channel;
3) \textbf{FF-K} replaces the kernel-based representation learning with a GNN-based aggregation scheme, i.e., replacing $\hat{\mathbf{z}}_{q, v}$ and $\hat{\mathbf{x}}_{q, v}$ with $\mathbf{z}_{q}$ and $\mathbf{x}_{q}$ in Eqs.~(\ref{eqn:kgat_textual_1})(\ref{eqn:kgat_user_interact}); 
4) \textbf{FF-I} excludes node importance learning 
and assigns an equal weight to every node; 
5) \textbf{FF-M} eliminates mutual-reinforcement-based evidence ranking, and selects the evidence for each topic $t$ by random sampling.
The ablation study results in Fig.~\ref{fig:ablation} show that removing each component leads to a decrease in model performance,
which demonstrates the effectiveness of our major components. 
A result of \textbf{FF-T}, which replaces our pretrained text encoder with that of KAN, can be found in the supplement.

We then conduct a sensitivity analysis of \emph{FinerFact} by changing the number of kernels $\tau$. 
Fig.~\ref{fig:sensitivity} shows that \emph{FinerFact} consistently outperforms the state-of-art baseline 
\emph{KAN} with varying numbers of kernels, 
which demonstrates the robustness of our method.
In addition, the performance is the best when using around 11 kernels.
Using more kernels does not necessarily lead to better performance due to overfitting.

\subsection{Case Study}

%


In addition to improving accuracy, our method also enables humans to understand most parts in the reasoning workflow. 
In this case study, we 
illustrate how \emph{FinerFact} reasons about the authenticity of a news story, which 
is about FBI lawyer Lisa Page disclosing that she was instructed to cover-up China's hacks of the DNC server. 
\emph{FinerFact} successfully identifies that the news is fake, with a detailed explanation about the salient evidence, subtle clues, and the prediction scores for each viewpoint.


\textbf{Identifying salient evidence.}
As shown in Fig.~\ref{fig:case_study}(a), \emph{FinerFact} identifies meaningful and relevant keywords that belong to diverse topics.
For each topic $t$, we can further understand its key evidence by observing its salient posts $P^t$ and average user saliency $\bar{r}_{U^t}$.
As shown in Fig.~\ref{fig:case_study}(b), users of topic 1 support the news because of their political stance (\xiting{consider China as a threat}) and are generally not credible opinion leaders (small $\bar{r}_{U^t}$). 
In contrast, users of topic 4, who question that the news is fake with more objective reasons, e.g.,  \xiting{it is unlikely that an outsider lawyer knows about the server hack} (\xiting{Fig.~\ref{fig:case_study}A}), receive more attention (larger $\bar{r}_{U^t}$). 




\textbf{Reasoning with subtle clues}.
The token-level and user-level attention scores $\alpha^i_{q,v}$ and $\rho^i_{q,v}$ reveal the subtle clues \emph{FinerFact} detects.
For example, in topic 2, the words with the largest $\alpha^i_{2,1}$ and $\alpha^i_{2,4}$ are ``china" and ``conspiracy".
These clues are meaningful and interesting:
the statement about ``china" being unlikely to collaborate with FBI in topic 2 decreases the credibility of the posts in topic 1, and relating ``conspiracy" with topic 4 (about ``hacking")  enables us to understand that the news may be fake because such \xiting{a hacking conspiracy is likely to be made up by people who like to talk about it} (\xiting{Fig.~\ref{fig:case_study}B}).
Topics 1 and 4 are also related: user 2, who has the largest $\rho^i_{4,1}$, questions users in topic 1 by commenting on them, and points out that China's problem in terms of intellectual property does not mean that Chinese will hack the server (\xiting{Fig.~\ref{fig:case_study}C}).  

\textbf{Prediction for each viewpoint.}
Based on the subtle clues, \emph{FinerFact} 
makes prediction for each node.
Our method understands that evidence from groups 1 and 3 imply that the news is true ($\mathbb{P}(y=0|v,G)>0.5$) and that the evidence from groups 2 and 4 imply that the news is fake ($\mathbb{P}(y=0|v,G)<0.5$).
It assigns a low probability score that is close to 0.5 to group 1 by propagating the information from groups 2 and 4 to group 1 (large $\gamma_{q,v}$).
It also assigns a small node importance $\mathbb{P}(v|G,R)$ to group 1.
This is reasonable, since group 1 has a low user saliency $\bar{r}_{U^t}$, which can be modeled by using the attention prior.
While the users in group 3 are considered salient according to the mutual reinforcement graph, we find that they are not talking about whether the article is true, but are instead drifting towards criticizing Hillary.
Our model successfully identifies this and assigns a low node importance to topic 3.

\textbf{Steering the model.}
\emph{FinerFact} also provides opportunities for users to steer and refine the model.
For example, we may integrate \emph{FinerFact} with the method proposed by~\citet{liu2015uncertainty} to enable interactive refinement of evidence ranking. Please refer to the supplement for more details.\looseness=-1

%% file: src/related.tex
\section{Related Works}




Methods for fake news detection can be divided into two main categories: content-based and knowledge-aware.


\textbf{Content-based} methods mainly utilize the textual or visual content from the news article and related posts for news verification~\cite{yang2012automatic,afroz2012detecting,kwon2013prominent,przybyla2020capturing,ma2016detecting,zellers2019defending,qi2019exploiting,gupta2013faking, jin2016novel, kaliyar2021fakebert}.
These methods enable the detection of fake news at an early stage~\cite{wei2021towards, pelrine2021surprising}.
However, their performance is limited as they ignore auxiliary knowledge for news verification.


\textbf{Knowledge-aware} methods leverage auxiliary knowledge for news verification
~\cite{ruchansky2017csi,wang2018eann,shu2019defend,jin2016news,ma2018rumor,cho2014properties,wang2018dkn}. 

These methods typically utilize external knowledge about entity relationships~\cite{xu2022mining, dun2021kan, pan2018content, silva2021embracing, hu2021compare} or social knowledge about online posts~\cite{lu2020gcan, nguyen2020fang, khoo2020interpretable, bian2020rumor} for fake news detection.
While existing methods have demonstrated the usefulness of heterogeneous social relations and external information~\cite{yuan2019jointly},
they either do not model the interactions between the news content and different types of knowledge data, or model them at a coarse-grained (e.g., sentence or post) level, which limits their performance.
We tackle this issue by proposing a prior-aware bi-channel kernel graph network, which enables fine-grained reasoning and improves detection accuracy.


%% file: src/conclusion.tex
\section{Conclusion}
We propose \emph{FinerFact}, a fine-grained reasoning framework for explainable fake news detection. We devise a mutual-reinforcement-based method for efficient evidence ranking and a prior-aware bi-channel kernel graph network for fine-grained reasoning on multiple groups of evidence. 
\yiq{Experimental results show the effectiveness of our method.}


%% file: src/supplementary.tex
\section*{Supplementary Material}

\subsection{Steering the Model}

\emph{FinerFact} is designed to provide a basis for humans to steer the model and refine it.
Here, we introduce three possible ways to steer our model by integrating human knowledge.

\textbf{Incremental evidence ranking refinement.}
Incorporating MutualRanker~\cite{liu2015uncertainty} into our model enables \emph{FinerFact} to incrementally refine evidence ranking based on user needs or insights.
For example, if a user identifies an important piece of evidence, e.g., an informative post, a keyword that should be paid more attention to, or a user that s/he agrees with, s/he can quickly refine the model locally to retrieve more related evidence, which will then help improve fake news detection.
The user could also decrease the saliency of the evidence that is not interesting.
For example, in the case study provided in the paper, most posts that mention ``hillary" are not discussing about the news content, but are instead drifting from the main topics to criticizing Hillary.
After identifying this issue, the user could decrease the saliency of ``hillary", and this information will be propagated to other related evidence incrementally to enable model refinement.

\textbf{Interactive salient topic refinement.}
Our topic-aware method provides a chance for humans to refine evidence ranking at a coarser granularity (topic-level).
For example, users can quickly understand each topic and select the topics of interesting for fine-grained reasoning.
Users can also increase the relative importance for an interesting topic during fine-grained reasoning by using the attention prior.
In particular, we may increase the saliency of all posts and users in the topic when computing the attention prior $\delta(v, R)$ in Eq.~(19).
This automatically increases the node importance for the final prediction.
Topic mining could also be systematically refined by adding must-link or cannot-link constraints~\cite{basu2004active}, or using interactive methods like Utopian~\cite{choo2013utopian}.

\textbf{Mutual reinforcement evidence graph refinement.}
Our mutual reinforcement method is designed based on human knowledge about attribute saliency (evidence with which attribute is important) and topological saliency (how saliency should propagate among evidence).
Currently, we provide a default setting summarized from the literature, which shows good performance.
Users could also refine this process based on their own insights and needs.
For example, if a user has an additional set of data, e.g., the number of likes a post receives, s/he can integrate such data by changing the way we compute the attribute saliency $E_{P}$ for posts.
Or, if a user wishes to better consider the semantic similarity between posts, s/he can change the definition of $A_{PP}$ from cosine similarity between term frequency vectors of the posts to the cosine similarity between the BERT~\cite{devlin2019bert} representations of posts.

\subsection{Implementation Details}

\subsubsection{Dataset}
The statistics of the datasets are described in Table~\ref{tab:dataset}. We remove notations in the posts including 1) hashtags starting with '$\#$'; 2) mentioning notations that starts with '$@$'; and 3) website URLs.

\yiq{For node-level label prediction, there is no training objective, since no node- or evidence-level label annotations are provided in the datasets.}

\begin{table}[t]
\centering
\begin{tabular}{|c|ccc|} 
\hline
           & \# True & \# Fake & \#~ Total  \\ 
\hline
PolitiFact & 443     & 372     & 815        \\
GossipCop  & 4219    & 3393    & 7612       \\
\hline
\end{tabular}
\caption{Statistics of the datasets}
\label{tab:dataset}
\end{table}

\subsubsection{Hyperparameter Settings}
The damping factor $d$ for the mutual reinforcement graph is set to 0.85 by following \cite{brin1998anatomy}.
For the mutual reinforcement graph, we set each intra-graph balancing weight $\beta_{xx}=1$ and set each inter-graph balancing weight $\beta_{xy}$ to $0.5$, where $x \ne y$. We set the damping factor $d$ to 0.85 according to \cite{brin1998anatomy}.  The smoothing constants $\theta_P, \theta_K, \theta_U$ in the computation of attribute saliency are all set to $1e-3$. $N_K, N_T$ are set to 7, 5, respectively. 
For fine-grained reasoning, we use the Adam optimizer \cite{kingma2015adam} and a learning rate of 5e-5. We use a batch size of 8 with accumulation step of 8.
\yiq{$N_K, N_T, N_S, N_P, N_U$ are set to 7, 5, 3, 6, 32 respectively.} We use 11 kernels to extract multiple levels of interactions. One kernel with $\mu_{\tau}=1$ and $\sigma_{\tau}=1e-3$ is used to model exact matches \cite{dai2018convolutional}. The other kernels all have $\sigma_{\tau}=0.1$, and their mean $\mu_{\tau}$ are evenly spaced within $[-1, 1]$. Each node $v$ contains 3 news sentences as its claim in $S_t$, 6 posts as its evidence in $P_t$. $N_U$ is set to 32.





\subsubsection{Mutual Reinforcement Evidence Graph Building}
The mutual reinforcement evidence graph $M=\{A_{xy}|x,y\in\{P,U,K\}\}$ is a three-layer graph, where subscript indices $P$, $U$, $K$ denote posts, users, and keywords, respectively, and $A_{xy}$ is an affinity matrix.
The edge weight $a_{p_i,p_j}$ between \textbf{posts} $p_i$ and $p_j$ is the cosine similarity between their term frequency vectors, which captures the in-between interactions between textually similar tweets~\cite{zangerle2013impact}.
\textbf{Users} are connected with commenting relationships, which  encodes the opinion diffusion within the social network.  
Specifically, $a_{u_{i}, u_{j}}$ is 1 if $u_i$ comments on $u_j$'s post and is 0 otherwise. 
Edges between the \textbf{keywords} are 
We consider two types of connections between \textbf{different types of evidence}:
1) mentioning relationship: a keyword is linked to all the posts and users that mentioned it; and
2) authoring relationship: a user is linked to all the posts that  s/he has published.
All edges except for the ones between the posts have a weight of 1.

Our idea is to use human knowledge to decide how the importance of each evidence should propagate.
For example, we can link users according to retweet relationships because a user that is retweeted by an important user is usually also important.
In comparison, linking users according to the order that they tweeted is less meaningful, as it is hard to say whether users who discussed the news earlier or later are more important.
Other candidate user relations such as mentioning the same words are reasonable, but they reveal user importance in a less direct way compared with retweets.


\subsection{Supplementary Experiments}

\subsubsection{Evaluation of Topic Quality}
\yiq{We evaluate whether topics are meaningful with case by case manual investigation (e.g., case study in Fig. 5). 
Our preliminary study with 5 users shows that like the topics in Fig. 5, most topics (51 out of 60) can be interpreted ($\geq$3 for a 4-point Likert scale) after checking their keywords and claims.}

\subsubsection{Leveraging Different Text Encoders}
To evaluate whether our method performs better than KAN when they use the same text encoder,
\yiq{
we implement \textbf{FF-T}.
This variant of our method replaces BERT in FinerFact with a 1-layer Transformer encoder, same as the text encoder used in KAN. 
Results show that \textbf{FF-T} performs better than KAN, with a Pre / Rec / F1 / Acc / AUC of 0.9091 / 0.9182 / 0.8810 / 0.8864 / 0.9234 and 0.8579 / 0.8874 / 0.8602 / 0.8310 / 0.8600 on PolitiFact and GossipCop, respectively.
We use randomly initialized word embedding in FF-T.
Leveraging word2vec and external knowledge about entities like that in KAN may further improve the result.
}

\subsubsection{Complexity and Efficiency}
\yiq{The complexity of our model is $O(N_v^2N_E^2)$, where $N_v$ is the maximum sequence length of a piece of evidence, and $N_E$ is the word embedding size. It takes 11 and 126 minutes to train FinerFact on PolitiFact and GossipCop, respectively.
}
\subsection{Discussion and Future Work}

\subsubsection{Evaluation for Interpretability}
The limitation of our work mainly pertains to the difficulty in evaluating interpretability.
Evaluating interpretability is a challenging and very important problem due to the lack of ground-truth.
Currently, we follow existing works~\cite{shu2019defend, lu2020gcan} to present a case study.
We are inspired to study more carefully about this in the future for helping the community, e.g., by collecting human annotations and conducting user studies.

\subsubsection{Using LDA for Topic Modeling}
LDA is not the best method for mining topics from
short text, and changing it to advanced methods like 
Twitter-LDA~\cite{zhao2011comparing} or PTM~\cite{zuo2016topic}
is likely to improve the interpretability of our method.
We would like to verify the effectiveness of Twitter-LDA, PTM or other topic mining methods on our datasets in the future.

\subsubsection{Relation with Feature Engineering}
The attribute saliency part of the method looks like feature engineering, which is an old technique that is gradually replaced by deep learning.
About this,
\yiq{we would like to consider our method as an
initial attempt to combine the benefits of traditional machine
learning (e.g., feature engineering and mutual reinforcement
graph) and deep learning. Traditional machine learning is
fast, predictable, requires much less labels, and could incorporate
human knowledge summarized based on years of
experience and careful analysis of the dataset. Meanwhile,
deep learning is more powerful, easy to implement, and can
identify complex patterns hidden in large-scale datasets. To
combine them, we put traditional machine learning in the
first module, which reduces the burden of deep learning by
efficiently identifying candidate evidence. We do not need
this part to be very accurate: it is okay that the retrieved candidates
contain noise. We then use deep learning in the second
module to perform accurate, data-driven, fine-grained
reasoning, which is very difficult for traditional machine
learning to achieve.}
We would like to explore more of such combinations in the future.

